\documentclass[anon]{l4dc2024}
\usepackage[]{xcolor}
\usepackage{algorithm, algpseudocode}
\usepackage{sidecap}
\usepackage{wrapfig}
\usepackage{amssymb}

\newcommand{\statespace}{\mathcal{S}}
\newcommand{\actionspace}{\mathcal{A}}
\newcommand{\propertyspace}{\mathcal{M}}
\newcommand{\observationspace}{\Omega}

\usepackage{graphicx}

\long\def\invis#1{}

\usepackage{float}
\newcounter{gdTmp}
\newcounter{gdLastCount}
\newcommand\maxpage[2][Error]{  
    \ifnum\value{page}>#2
        \gderror{{\Large On page {\thepage} we are past page #2 (too long).   #1 }}
    \else\fi
    \setcounter{gdLastCount}{\value{page}} 
}
\newcommand\maxpageSinceLast[2][Error]{  
    \ifnum \numexpr \value{page} - \value{gdLastCount}\relax>#2
        \gderror{Exceeds max length #2 pages. Page \thepage: #1}
        \thepage\else\fi
    \setcounter{gdLastCount}{\value{page}}
}

\usepackage{amsfonts}
\usepackage{graphicx}
\usepackage{amsmath}

\usepackage{acronym}

\usepackage{hyperref}

\acrodef{POMDP}{partially observable Markov decision process}
\acrodefplural{POMDP}{partially observable Markov decision processes}
\acrodef{MDP}{Markov decision process}
\acrodef{EKF}{extended Kalman filter}
\acrodef{UKF}{unscented Kalman filter}
\acrodef{HMM}{hidden Markov model}
\acrodefplural{HMM}{hidden Markov models}
\acrodef{IID}{independent and identically distributed}
\acrodef{VAE}{Variational autoencoder}
\acrodef{RNN}{recurrent neural network}
\acrodef{GRU}{gated recurrent unit}
\acrodef{MAE}{mean absolute error}
\acrodef{ELBO}{evidence lower bound}
\acrodef{MPC}{model-predictive control}
\acrodef{COM}{center of mass}
\acrodef{IMU}{inertial measurement unit}
\acrodef{PO-UCT}{partially observable upper confidence bound applied to trees}

\title[Active tactile perception]{Learning active tactile perception through belief-space control}
\usepackage{times}

\author{%
 \Name{Jean-Fran\c{c}ois Tremblay} \Email{jft@cim.mcgill.ca}\\
 \addr Center for Intelligent Machines, McGill University, Canada
 \AND
 \Name{David Meger} \\
 \addr Center for Intelligent Machines, McGill University, Canada
 \AND
 \Name{Francois Hogan} \\
 \addr Samsung AI Center Montreal, Canada
 \AND
 \Name{Gregory Dudek} \\
 \addr Center for Intelligent Machines, McGill University \& Samsung AI Center Montreal, Canada
}

\begin{document}

\maketitle

\begin{abstract}
    Robots operating in an open world will encounter novel objects with unknown physical properties, such as mass, friction, or size.
    These robots will need to sense these properties through interaction prior to performing downstream tasks with the objects.
    We propose a method that autonomously learns tactile exploration policies by developing a generative world model that is leveraged to 1) estimate the object's physical parameters using a differentiable Bayesian filtering algorithm and 2) develop an exploration policy using an information-gathering model predictive controller.
    We evaluate our method on three simulated tasks where the goal is to estimate a desired object property (mass, height or toppling height) through physical interaction.
    We find that our method is able to discover policies that efficiently gather information about the desired property in an intuitive manner.
    Finally, we validate our method on a real robot system for the height estimation task, where our method is able to successfully learn and execute an information-gathering policy from scratch.
\end{abstract}

\section{Introduction}

This paper proposes an active perception framework that can recover the dynamical properties of objects by manipulating them, without prior knowledge of their geometry or inertia.
Akin to active vision \citep{activevision}, we seek an approach that can drive a robot to move in an information-seeking manner.
Unlike shape or object class, the dynamical object properties we recover are not fully observable from a single  sensor reading and require sequential and physical interaction.

Psychology literature refers to the way humans measure these properties as \textit{exploratory procedures} \citep{LEDERMAN1987342}.
These procedures are shared amongst human subjects and hint to a common understanding of how to effectively manipulate objects to gather their information.
For example, object hardness is commonly perceived by \textit{pressing} down with our fingers while object mass is typically estimated by \textit{lifting} vertically. 
These exploratory procedures are challenging to hand-engineer in the general case and vary largely based on the object class. In this paper, our objective is to develop an active perception framework that can autonomously generate useful robotic exploratory procedures.

We propose a method that autonomously learns tactile exploration policies by developing a generative world model that is leveraged to
1) estimate the object’s physical parameters using a differentiable filtering algorithm, leveraging a novel way to train a generative world-model end-to-end and
2) develop an exploration policy using an information-gathering model predictive controller.



The first aspect of our method involves learning a model that can infer the dynamical properties from a sequence of force and proprioceptive measurements, where the underlying physical motions include contact changes, multi-body friction systems and the unknown dynamics of a 7-DoF robot.
To achieve this, we derive a novel loss for learning-based generative sequence modeling leveraging differentiable Kalman filters.
We consider the task of recovering object properties, such as mass and height.
While the object properties are provided during the training phase, they are unknown  during deployment, at which time our algorithm infers the properties of unseen objects.
Our learned generative model allows to simulate the belief process forward in time and answers the question: ``What will the uncertainty about my state be in the future?''.
This enables us to select actions such that they minimize the uncertainty about the object property.

The second aspect of our method is an information seeking controller that plans trajectories specifically for each task and object type.
Our method produces diverse motions targeted to each desired property.
Notably, this is  done without  human engineering of the appropriate motions, unlike many past works \citep{swingbot, denil2016learning}.
Our policy begins from random initial motions that provide only limited contact. The robot progressively gathers experience to refine the generative model, which in turn leads to more informative contacts at each iteration.

The contributions of our approach are as follows:
\begin{itemize}
    \item The derivation and implementation of a novel generative filtering model that learns 1) the dynamics of the systems and 2) the generative observation model (without state supervision, as opposed to previous work \citep{bohgukf}).
    \item An information-guided active manipulation system capable of generating diverse exploratory procedures; and
    \item Experiment validation in simulation and on a real-world task, where we find that our method yields better data efficiency and estimation error than a Deep Reinforcement Learning baseline.
\end{itemize}


After first exploring related works in \autoref{sec:related}, we discuss in \autoref{sec:methods} the theoretical setting for the problem, describe our model and derive the novel loss function and procedure for training and using the resulting model to control the robot.
We describe the experimental setup for simulation and the real system in \autoref{sec:experiments} and finally show the results in \autoref{sec:results}.


\section{Related works}
\label{sec:related}


\paragraph{State estimation}
has been used to estimate properties of systems.
For example, \citet{imudrift} estimate the bias parameters of an \acf{IMU} in the context of localization and mapping.
\citet{stergios} used a Kalman filtering approach to do \ac{IMU}-camera calibration.
There are several works proposing the fusion of Bayesian filtering methods with deep learning where the dynamics and observation models are  neural networks.
\citet{bohgukf} provide a good overview of learning Bayesian filtering models for robotics applications, and release \texttt{torchfilter}, a library of algorithms for this purpose which we build on for our belief-space control algorithm.  While this method can learn generative observation models, they require full state supervision during training. Here, we will present a new way to learn a generative world model using Kalman filtering that only requires noisy observations and an object property of interest.
In \citep{backpropkf}, the authors present the Backprop Kalman filter  described as a discriminative approach to filtering.
Discriminative filtering replaces the observation model with a learned mapping from observation to state instead.
In this paper, we propose to learn a generative observation model, which is key to predicting future state uncertainty and planning for informative actions.

\paragraph{Active perception}
\citep{activeperception} consists of acting in a way that assists perception and can incorporate learning, including the learning methods above.
\citet{denil2016learning} use reinforcement learning for ``Which is Heavier'' and ``Tower'' environments, where the goal of the former is to push blocks and, after a certain interaction period, take a ``labelling action'' to guess which block is heavier.
They train a recurrent deep reinforcement learning policy on that environment.
The action space for these problems is constrained and designed to act such that the blocks are pushed with a fixed force towards their center of mass. While this method enables the robot to effectively retrieve mass using human priors and intuition, our work differs where the robot is tasked with discovering such behaviors autonomously with unconstrained action spaces.
\citet{swingbot} introduce SwingBot, a robotic system that swings up an object with changing physical properties (moments, center of mass).
Before the swing up phase, the system follows a hand-engineered exploratory procedure that shakes and tilts the object in the hand to extract the necessary information for a successful swing up. Rather than engineering the exploration phase, we propose a generic framework for extracting such information before accomplishing a given task.

\paragraph{Model-based reinforcement learning}
approaches \citep{dyna, planet, pilco} learn a dynamics model for the system and use it to perform control (either through a sample based controller or by leveraging the model to learn a value function), similarly to our method.
However, they typically do not predict the future beliefs about the state of the system and thus, cannot plan to gather information.

\section{Methods}
\label{sec:methods}
This section outlines our approach to active tactile perception that   1) estimates the current object property from past observations,  2) simulates future observations given a sequence of  actions, 3) estimates future state uncertainty from those future observations, and 4) controls the robot arm to effectively recover the object property of interest.

\begin{figure}
    \centering
    \includegraphics[width=0.6\columnwidth]{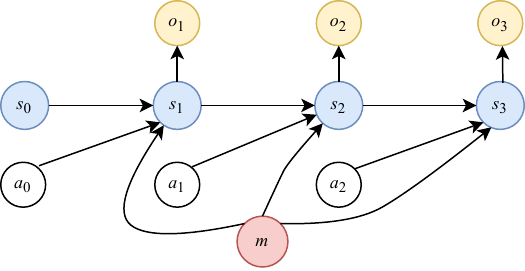}
    \caption{The dynamical Bayesian network representing the structure of our problem. $s_t$ represents the hidden state, $m$ represents the property of interest, $a_t$ and $o_t$ represent the actions and observations.
    }
    \label{fig:bayesnet}
\end{figure}


\begin{sloppypar}
    Let us begin with the theoretical framework in which we will formulate this problem.
    We consider a \acf{POMDP} setting, where each  observation $o_t$ gives us partial information about the state of the robot and object we are interested in.
    This formulation also includes an additional object property $m \in \mathcal{M}$, as illustrated in \autoref{fig:bayesnet}.
    More formally this \ac{POMDP} is a tuple $(\statespace, \mathcal{M},  \actionspace, p(s_{t+1}|s_t, m, a_t), \Omega, p(o_t|s_t), c)$, where
    the state, property, action and observation space ($\statespace$, $\propertyspace$, $\actionspace$ and $\observationspace$ respectively) are in $\mathbb{R}^n,  \mathbb{R},  \mathbb{R}^m, \mathbb{R}^d$ respectively.
    $c$ represents the cost function for the \ac{POMDP}.
    The representation for the state will be learned in a self-supervised fashion, as described in \autoref{sec:filtering}.
\end{sloppypar}

We are in an episodic setting with ending timestep $T$, and where at each episode the object is randomized.
Our aim is to learn to jointly estimate the state $s_t$ as well as the property $m$ using bayesian filtering, by  appending the property to the state and running the bayesian filter on the augmented state space $\statespace \times \mathcal{M}$.
This approach is analogous to EKF-SLAM \citep{ekfslam} where we jointly estimate the state of the robot as well as the pose of the landmarks in the environment, or system parameter estimation methods from the stochastic control literature where we jointly estimate the state of the system and some of its parameters \citep[][\S 4.7]{stengel1994optimal}. However, unlike the aforementioned methods,  the model is unknown and we must jointly estimate the property and  learn the dynamics and observation models.

\begin{sloppypar}
    In \autoref{sec:filtering} we  describe how to learn a model which infers the belief state (containing an estimate of the object property of interest)
    $b_t \approx p(s_t, m|a_0, \dots, a_{t-1}, o_1, \dots, o_t)$
    and $\bar{b}_t \approx p(s_t, m|a_0, \dots, a_{t-1}, o_1, \dots, o_{t-1})$ the marginals for the property or state being written as $b_t^m$, $b_t^s$, $\bar{b}_t^m$, $\bar{b}_t^s$. In \autoref{sec:control} we use this model to design an information-gathering controller.
    Finally, in \autoref{sec:loop} we present how to integrate these two things in a data-collection/training and control loop.
\end{sloppypar}

In our formulation, the cost $c$ is defined on the belief-space as $c(b_t)$ rather than the state-space $c(s_t)$ \citep{rhopomdp}.
This formalism allows one to penalize uncertainty about the state, which is crucial to minimizing uncertainty about the property of interest as will be done is \autoref{sec:control}.

\subsection{Learning-based Kalman filter}
\label{sec:filtering}
We develop a Kalman filter-based architecture where the objective is to learn a dynamics and observation model while performing belief-state inference, using the ground-truth property that will be available at training time.
In this section, we derive a loss function that combines the observation model and the property estimation.

The dynamics model representing $p(s_t|s_{t-1}, m, a_{t-1})$ is
\begin{equation}
    s_t = f_{\theta}(s_{t-1}, m, a_{t-1}) + \Sigma_{\theta}(s_{t-1}, m, a_{t-1})w_t
\end{equation}
where $w_t$ are \ac{IID} standard Gaussian random variable in $\mathbb{R}^n$,
$f_\theta: \statespace \times \propertyspace \times \actionspace \rightarrow \statespace$ and
$\Sigma_\theta: \statespace \times \propertyspace \times \actionspace \rightarrow \mathbb{S}^n_+$ with $\mathbb{S}^n_+$ being the space of positive-definite symmetric matrices of size $n$.

The dynamics model of the property-estimating filter can be written in the augmented state-space as:
\begin{equation}
    \begin{bmatrix}
        s_t \\
        m_t
    \end{bmatrix} =
    \begin{bmatrix}
        f_{\theta}(s_{t-1}, m_{t-1}, a_{t-1}) \\
        m_{t-1}
    \end{bmatrix} +
    \begin{bmatrix}
        \Sigma_{\theta}(s_{t-1}, m_{t-1}, a_{t-1}) & \mathbf{0}  \\
        \mathbf{0}^T                               & \varepsilon
    \end{bmatrix}\hat{w}_t
\end{equation}
with $\hat{w}_t$ being a standard normal random variable in $\mathbb{R}^{n+1}$ and
$\varepsilon$ is a small constant so that the covariance matrix remains definite-positive for numerical computation reasons.

Generative filtering (as opposed to discriminative filtering \citep{bohgukf, backpropkf, dkf}) implies learning a generative world-model, able to fully simulate the system and generate observations via the equation
\begin{equation}
    o_t = h_{\theta}(s_t) + \Gamma_{\theta}(s_t)v_t
\end{equation}
where $v_t$ are \ac{IID} standard Gaussian random variables in $\mathbb{R}^d$,
$h_\theta: \statespace \rightarrow \observationspace$ and
$\Gamma_\theta: \statespace \rightarrow \mathbb{S}^d_+$.
$\theta$ is the parameters for the neural networks $\Sigma, h, \Gamma$, which in this work are multilayer perceptrons with residual connections, and $f$ which is a \ac{GRU}.
The covariance networks for $\Sigma_\theta$ and $\Gamma_\theta$ output the diagonal of the square root of the covariance matrix (in the Cholesky sense), with the off-diagonal elements being 0 in our case.
Next is the derivation of the loss function for our model, which has two terms: one to train the observation and dynamics model in a self-supervised manner, the other to train our model to estimate the object property.

\paragraph{Observation loss}
Using an explicit-likelihood setting, we train the model in an self-predictive manner.
In \autoref{eq:generative_observation}, we present the derivation for the loss of the generative observation model.
This derivation is adapted and extended from \citep[][\S 12.1.1]{sarkka2013bayesian}, where we include action variables.
\begin{align}
    p(o_1, \dots, o_T | \theta, a_0, \dots, a_{T-1}) & = \prod_{t=1}^{T} p(o_t|\theta, o_1, \dots, o_{t-1}, a_0, \dots, a_{t-1})                                           \\
                                                     & = \prod_{t=1}^{T} \int_{\mathbb{R}^n} p(o_t|\theta, s_t)p(s_t|\theta, o_1, \dots, o_{t-1}, a_0, \dots, a_{t-1})ds_t \\
                                                     & \approx \prod_{t=1}^{T} \int_{\mathbb{R}^n} p(o_t|\theta, s_t)\bar{b}_t^s(s_t|\theta)ds_t                           \\
                                                     & = \prod_{t=1}^{T} \mathbf{E}_{s_t \sim \bar{b}_t^s(s_t|\theta)} p(o_t|\theta, s_t)
    \label{eq:generative_observation}
\end{align}

If we take the log, get a lower bound from Jensen's inequality and compute the empirical mean, we get:
\begin{align}
    \label{eq:elbo}
    \log p(o_1, \dots, o_T | \theta, a_0, \dots, a_{T-1}) & \gtrapprox \sum_{t=1}^{T} \frac{1}{N}\sum_{i=1}^{N} \log p(o_t|\theta, s_t^i) \hspace{15pt} s_t^i \sim \bar{b_t^s}(s_t|\theta) \\
                                                          & := \mathcal{L}_o
\end{align}
\autoref{eq:elbo} gives us a novel lower bound of the log likelihood (similarly to the ELBO loss in VAEs \cite{vae}) to train our model leveraging the differentiable \ac{EKF} \citep{bohgukf} used to compute $\bar{b_t^s}$.
Because $\bar{b_t^s} = \mathcal{N}(s_t|\bar{\mu_t}, \bar{\Sigma_t})$, we can use the reparametrization trick to sample $s_t^i$ by sampling $\xi^i$ from a $n$-dimensional standard Gaussian, and then letting
$
    s_t^i = \bar{\mu_t} + \bar{\Sigma_t}\xi^i
$

\paragraph{Object property loss}
Additionally to the ability to generate observations, we want our model to be able to estimate the object's property of interest.
To achieve this, we maximize the likelihood of the ground-truth property which is known at training time:
\begin{equation}
    \mathcal{L}_m = -\sum_{t=1}^T\log b_t^m(m)
\end{equation}
Where $b_t^m$ is a Gaussian density, given by our \ac{EKF} which estimates the state and the property of interest.

The loss we minimize is a combination of the self-predictive loss for the observation, and the likelihood of the mass in the state representation with $\alpha$ as a weighting parameter for the two losses.

In practice, we sample batches of sequences of length less than $T$, and initialize the filter using stored beliefs in the dataset, in a truncated backpropagation through time fashion~\citep{tbptt}.

\subsection{Information-gathering model-predictive controller}
\label{sec:control}

\begin{figure}
    \centering
    \includegraphics[width=0.7\columnwidth]{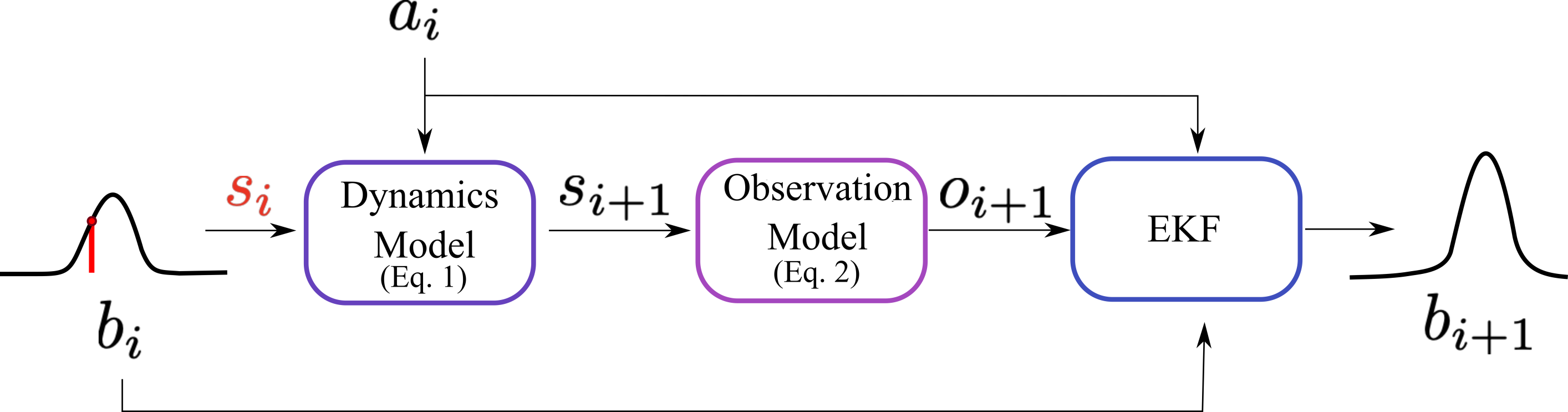}
    \caption{Illustration of the sampling process for belief dynamics using a generative model.
        First, states are sampled from the current belief.
        We can then use our dynamics model and candidate actions to sample future states.
        These future states are given to our generative observation model to generate observations.
        We can then feed the generated observations and candidate actions to the state estimator to simulate the belief-space dynamics.}
    \label{fig:sampling}
\end{figure}

The goal is to control the belief space process in a way that collects information about the property we're trying to perceive.
We describe the belief dynamics, cost function and optimizer necessary to achieve this.

\paragraph{Belief dynamics}
We can use the learned world model to simulate the belief space dynamics, as illustrated in \autoref{fig:sampling}.
The key is to be able to use the learned observation model to predict the future uncertainty about the state, rather than merely predict future states.

\paragraph{Cost function}
We want our controller to minimize the entropy $H$ of the belief of the object property, giving us the cost
$
    J = \sum_{t=1}^T H(b_t^m)
$
to minimize the uncertainty about the property of the object as soon as possible in the episode.
Minimizing this cost, for a Gaussian belief $b_t^m$ with mean and variance $\mu_t^m, \sigma_t^m$, is equivalent to minimizing the cost
\begin{equation}
    \label{eq:cost}
    J = \sum_{t=1}^T \log \sigma_t^m.
\end{equation}

\paragraph{Optimizer}
We used a sampling-based optimizer which selected the best randomly-generated sequence of actions, minimizing the cost of \autoref{eq:cost}.
The actions are generated using a Gaussian random walk in $\mathbb{R}^3$, with a standard deviation of $10$ cm for all tasks.
Following the \ac{MPC} framework, we execute the first action of the sequence and then re-optimize.



\subsection{Full training and control loop}
\label{sec:loop}

During training, we follow the  procedure:
1) Collect data using current controller for one epoch (randomizing the object property of interest), saving the observations, actions and estimated beliefs as well as the ground truth object property for this epoch
2) Train the state estimator using the dataset
3) Update stored beliefs in the dataset (by replaying the actions and observations).

Step 3) does not have to be done every epoch and can be costly as the dataset grows, but it is important to perform truncated backpropagation through time and initialize our state estimate during training.

\subsection{Deep reinforcement learning baseline}
\label{sec:baseline}
We compare our method to a model-free deep reinforcement learning approach.
We follow the method used by \citet{denil2016learning} and augment the action space with an estimate of the object property, that is, at every timestep the agent both acts and makes a prediction about the property of interest, getting a reward based on how good the prediction is.
The main difference with the formulation of \citet{denil2016learning} is that we make a prediction at every time step rather than  at the end of the exploration period. This is to be consistent with our belief-space control cost (\autoref{eq:cost}) where we constantly evaluate the accuracy of the estimate rather than only judging the final estimate.
More formally, the agent has an augmented action $\bar{a}_t = (a_t, \mu_t, \sigma_t)$ and gets a reward $r_t = \mathcal{N}(m|\mu_t, \sigma_t)$ where $m$ is the ground-truth object property.
To account for the partially observable nature of the task, we stack the past $30$ observations and encode them with a fully-connected neural network (we tried a \ac{GRU} \citep{cho2014properties} with no benefit for this medium-horizon task).
We use a TD3 \citep{td3} based agent with the augmented action space.

\section{Experiments}
\label{sec:experiments}

In this section, we validate our proposed active perception framework introduced in \autoref{sec:methods} in both  simulation and a real robot experiments.
We set up three custom robosuite \citep{robosuite2020} environments for our experiments.
For all experiments, we use a Franka Emika arm, as shown in \autoref{fig:demo_weight} and \autoref{fig:demo_real}, and a force-torque sensor at the wrist.
We use impedance position control in three degrees of freedom, sending delta translation commands to the robot.
The observations consist of the robot joint poses as well as the force and torque measurements at the wrist.

In all experiments, the dynamics model $f_\theta$ is a \acf{GRU} \citep{cho2014properties}, with a state-space dimension of 128.
The observation model $h_\theta$ is a fully connected five layer neural network, each hidden layer having 128 units.
The two noise models $\Sigma_\theta$ and $\Gamma_\theta$ use the same fully-connected architecture, outputting the diagonal of the covariance (outputting the full Cholesky decomposition of the covariance was found to have no benefit).

\paragraph{Mass estimation in simulation} The first task is to learn to estimate the mass of a cube. The cube has a constant size and coefficient of friction, but its mass changes randomly between $1$ kg and $2$ kg  between episodes. Because the robot is equipped with a single finger, it cannot grasp the object vertically. An expected behavior would be  to push it forward to extract mass information from the force and torque readings.

\paragraph{Height estimation in simulation}
The second task is to learn to estimate the height of a block, randomized between $1$ cm and $15$ cm.
In this scenario. the force torque sensor also behaves as a contact detector. An expected behavior would be to poke the object from above, at which point the height could be extracted from forward kinematics (unknown to the robot). One subtlety is that the arm must position itself above the box prior to moving down, as it can otherwise make contact with the table instead.

\paragraph{Toppling height estimation in simulation}
The third task consists is to estimate the minimum toppling height of an object. That is the height above which the object will topple instead of slide when pushed.
The object is L-shaped, with a variable feet length and mass which influences the toppling height. An expected behavior would be for the robot to tap the object forward at different vertical locations to detect if object pivots or slides forward.

\paragraph{Height estimation on a real system}
We validate our approach on a real-robot robotic experiment for the height estimation task described previously.
The setup mimics the height estimation task described above, where an actuated platform changes the  height of the object relative to the table between each episode uniformly between $27$ cm and $34$ cm.
We learn the policy from scratch on the real system without use of any transfer learning.

\section{Results}
\label{sec:results}
\subsection{Simulation}

\begin{figure}
    \centering
    \includegraphics[width=\columnwidth]{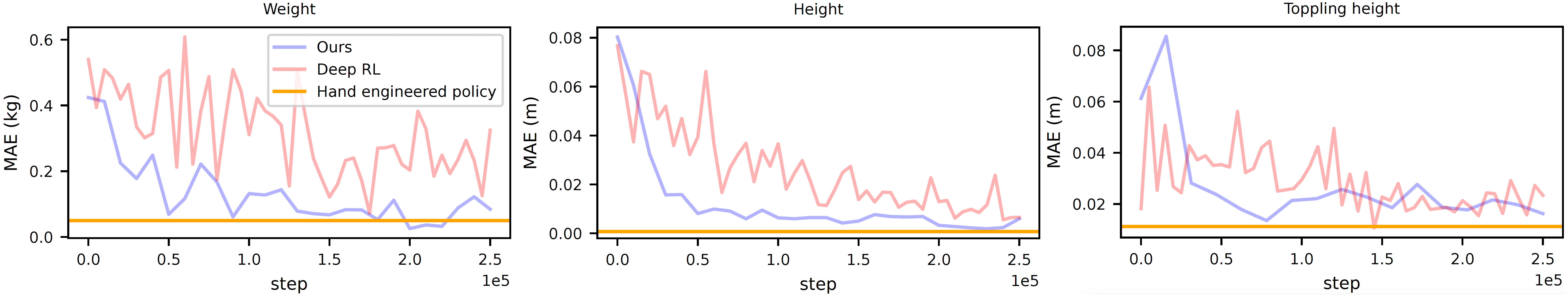}
    \caption{\Acf{MAE} for the property estimation tasks, at the end of the episode averaged over 10 runs, as learning progresses. The hand engineered policy gives an idea on what can be achieved when the behavior must not be discovered, and we simply have to extract the property from a sequence of sensor readings.}
    \label{fig:logprob}
\end{figure}

\begin{figure}[t!]
    \centering
    \begin{minipage}{0.49\textwidth}
        \vspace{-10pt}
        \centering
        \includegraphics[width=\columnwidth]{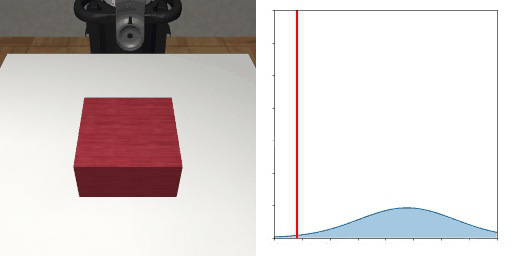} \\
        \includegraphics[width=\columnwidth]{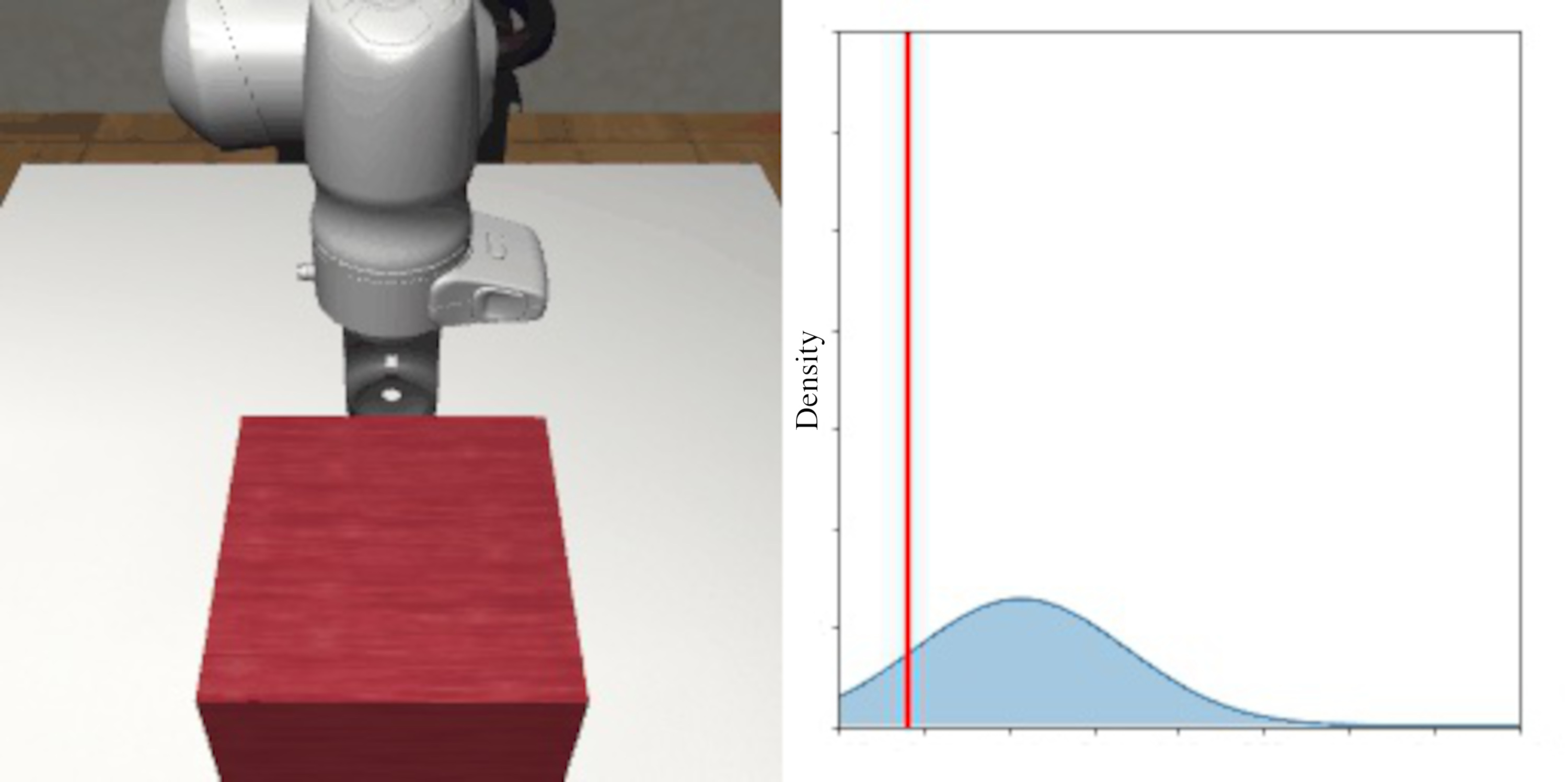} \\
        \includegraphics[width=\columnwidth]{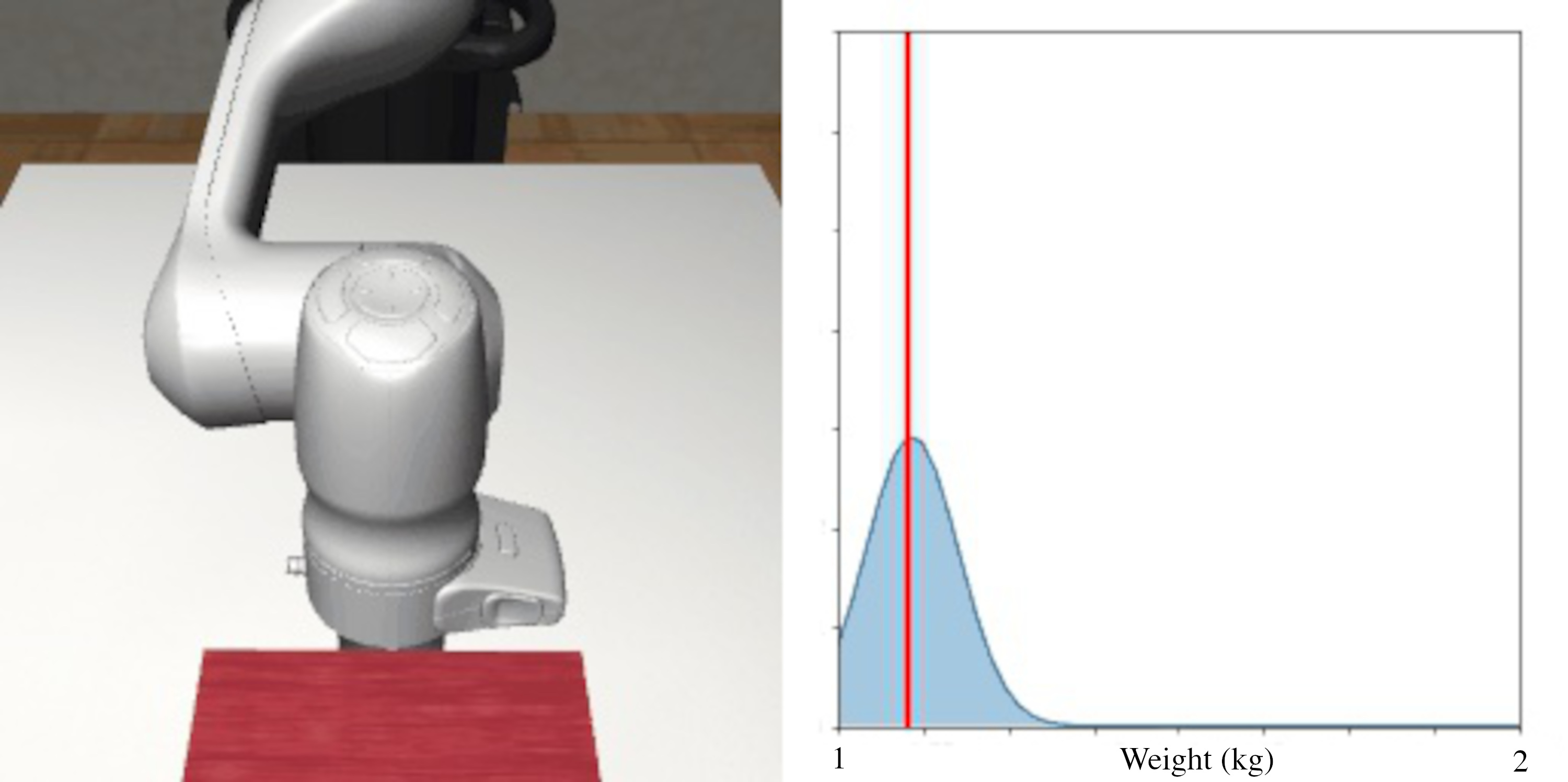}
        \caption{\vspace{-15pt} \textbf{Pushing interaction to estimate object mass}. We can see that the robot learns to stably push the object to extract mass from force torque readings. Notice how the uncertainty goes down as the arm starts pushing the block.}
        \label{fig:demo_weight}
    \end{minipage}
    \begin{minipage}{0.49\textwidth}
        \centering
        \vspace{-10pt}
        \includegraphics[width=\columnwidth]{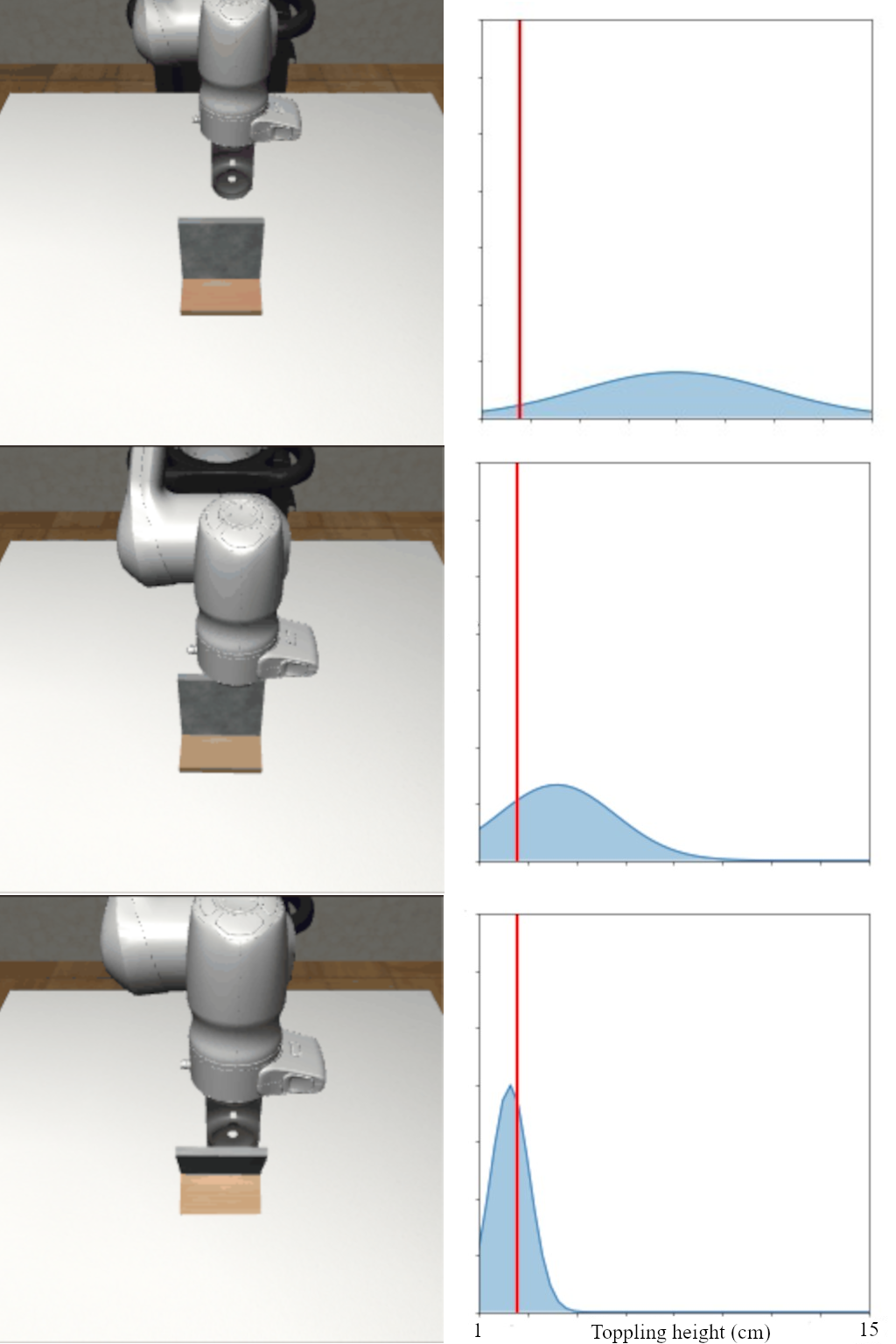} \\
        \caption{\textbf{Probing to estimate toppling height}. We can see that the robot learns to probe the object, starting from the bottom and gradually pushing while moving upwards, trying to topple the object.}
        \label{fig:demo_toppling_height}
    \end{minipage}
\end{figure}

For every simulated task, we evaluate the policy after every  $10000$ environment steps. The evaluation procedure runs $10$ episodes with  a randomized object property and computes the \acf{MAE}  using the estimate at the last timestep of the episode.
The training curve for each simulated task is shown in \autoref{fig:logprob} and shows the evolution of the \ac{MAE} during training.
In each result, we include the additional baseline of a hand engineering policy, which we depict with a yellow  line. The line is horizontal as the policy remains fixed through time while the object property is extracted using the state estimator presented in \autoref{sec:filtering}. Note that our method, in all scenarios, tends towards a similar performance despite having the additional complexity of learning the exploratory procedure from scratch.
This baseline is included to provide a best-effort hand-engineered comparison for our information-gathering controller.
In the figure, we  also include the performance of the reinforcement learning baseline of \autoref{sec:baseline} in red.

We can see that as learning progresses, two things happen concurrently.
\textbf{First, the agent learns to perform informative actions.}
In mass estimation, the policy pushes the block stably as shown in \autoref{fig:demo_weight}.
In toppling height estimation, the policy pushes on the object at different height, starting from the bottom as seen in \autoref{fig:demo_toppling_height}.
In height estimation, the policy goes down in a straight line until it touches the blocks as shown in \autoref{fig:demo_real}.
\textbf{Second, the state estimator learns to extract the property from the observations generated by the informative actions.}
For example during height estimation, the uncertainty remains high until the end-effector touches the block, at which point the estimate peaks at the correct height.
It is important to note that the exploration strategies are in no way encoded in the agent. For example, the pushing strategy to recover mass is an emergent behavior learned by the agent from  initial random trajectories.

\subsection{Real system}
\begin{figure}
    \centering
    \includegraphics[width=0.24\columnwidth]{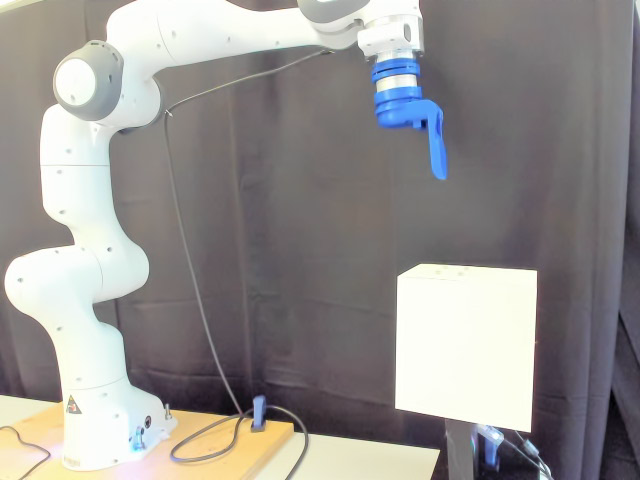}
    \includegraphics[width=0.24\columnwidth]{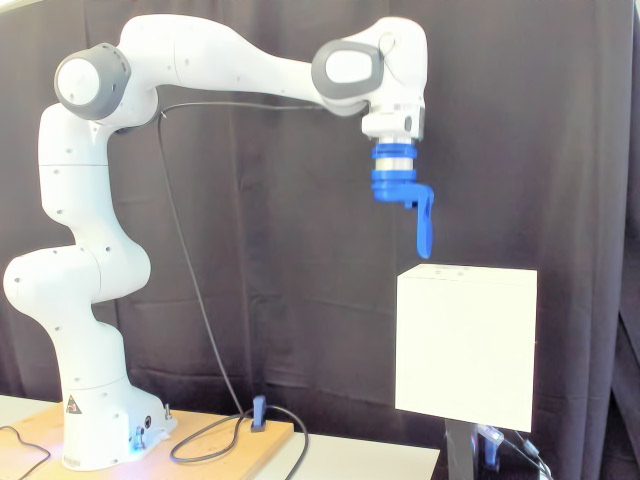}
    \includegraphics[width=0.24\columnwidth]{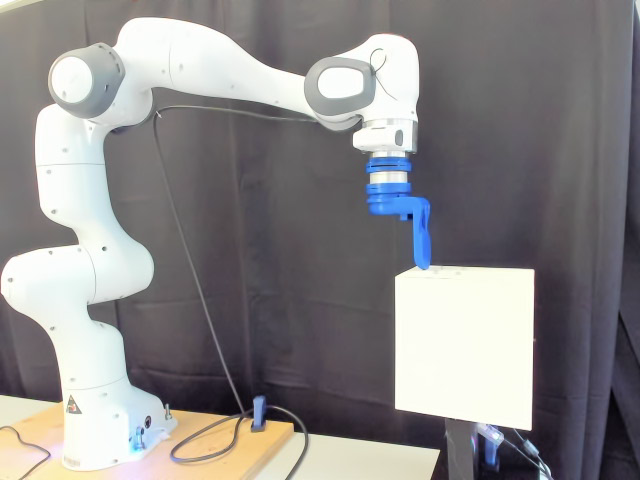}
    \includegraphics[width=0.24\columnwidth]{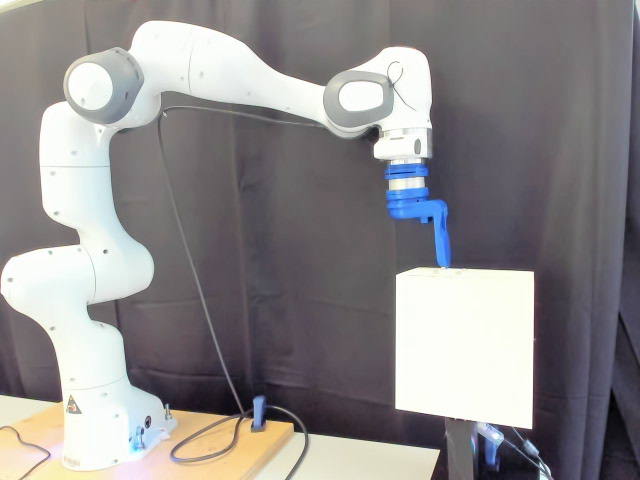} \\
    \includegraphics[width=0.24\columnwidth]{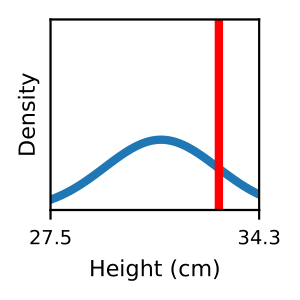}
    \includegraphics[width=0.24\columnwidth]{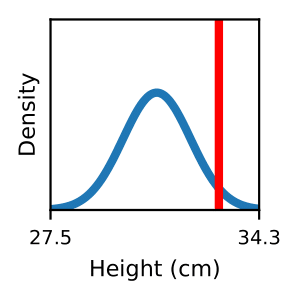}
    \includegraphics[width=0.24\columnwidth]{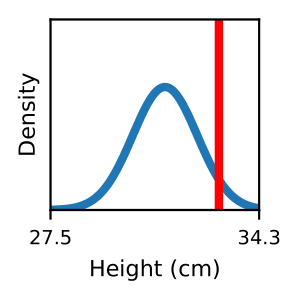}
    \includegraphics[width=0.24\columnwidth]{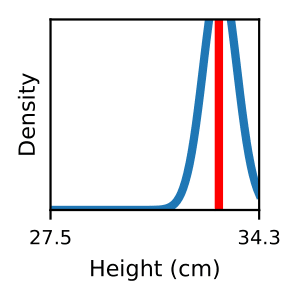}
    \caption{Demonstration of the learned controller for for real world height estimation. The robot correctly extracts the height of the platform, and the uncertainty goes down as the robot touches the robot for a longer period of time.}
    \label{fig:demo_real}
\end{figure}

A working example of the controller being deployed on the real robot is shown in \autoref{fig:demo_real}. The \ac{MAE} after $50 000$ timesteps is $1.19$ cm, with the range of motion of the platform being $7$ cm.
This is computed using $10$ evaluation episodes.

An important source of uncertainty is the height platform, which is only accurate within $5$ mm of its target height due to its flexible 3D printed nature and imperfect height controller driving the electric motor.
This real-world deployment also highlight the importance of the data efficiency of our approach, which succeeds after only eight hours of learning an end-to-end policy.

\section{Conclusion}
With the goal of discovering active tactile perception behaviors to measure object properties, we designed a novel active perception framework that includes a learning-based state estimator and an information-gathering controller.
Together, these two components allowed a robotic manipulation system to extract unknown object properties through physical exploration.
We validated our approach on three simulated tasks, where the robot was able to discover a pushing strategy for mass estimation, a top-down patting strategy for height estimation and a pushing strategy for toppling height estimation, without any prior on what should the trajectory be.
Furthermore, the approach was successfully deployed on a real-robot system for height estimation, demonstrating the ability of our approach to deal with the complexity of the real-world in a data-efficient manner.
This work opens up the door to learning more complex information-gathering policies, such as those for estimating the center of mass, hardness, friction coefficient and more.

\bibliography{bib.bib}
\end{document}